# Assessment of Sports Concussion in Female Athletes:
# A Role for Neuroinformatics?

Rachel Edelstein

Sterling Gutterman

Benjamin Newman

John Darrell Van Horn

Department of Psychology,

University of Virginia,

409 McCormick Road

Gilmer Hall room 304

Charlottesville, VA 22904

qbj6mv@virginia.edu

**Words in Abstract:** 237

**Words in Body:** 4,775

**References:** 122

## Abstract

Over the past decade, the intricacies of sports-related concussions among female athletes have become readily apparent. Traditional clinical methods for diagnosing concussions suffer limitations when applied to female athletes, often failing to capture subtle changes in brain structure and function. Advanced neuroinformatics techniques and machine learning models have become invaluable assets in this endeavor. While these technologies have been extensively employed in understanding concussion in male athletes, there remains a significant gap in our comprehension of their effectiveness for female athletes. With its remarkable data analysis capacity, machine learning offers a promising avenue to bridge this deficit. By harnessing the power of machine learning, researchers can link observed phenotypic neuroimaging data to sex-specific biological mechanisms, unraveling the mysteries of concussions in female athletes. Furthermore, embedding methods within machine learning enable examining brain architecture and its alterations beyond the conventional anatomical reference frame. In turn, allows researchers to gain deeper insights into the dynamics of concussions, treatment responses, and recovery processes. This paper endeavors to address the crucial issue of sex differences in multimodal neuroimaging experimental design and machine learning approaches within female athlete populations, ultimately ensuring that they receive the tailored care they require when facing the challenges of concussions. Through better data integration, feature identification, knowledge representation, validation, etc., neuroinformaticists, are ideally suited to bring clarity, context, and explainabilty to the study of sports-related head injuries in males and in females, and helping to define recovery.

**Statements and Declarations:**

The authors have no relevant financial or non-financial interests to disclose.

The authors have no conflicts of interest to declare that are relevant to the content of this article.

The authors don't have any research data outside the submitted manuscript file.

All authors certify that they have no affiliations with or involvement in any organization or entity with any financial interest or non-financial interest in the subject matter or materials discussed in this manuscript.

The authors have no financial or proprietary interests in any material discussed in this article.

**Introduction**

Historically, sports science research has tended to focus on sports involving primarily male athletes[1,7,9,10,11] where, with a traditionally larger pool of athletes to draw from, male participants have tended to be more accessible for research.[1,11] Indeed, until 1993[2] women were not required to be included in National Institutes of Health (NIH) supported clinical research studies. Moreover, societal attitudes have often reinforced gender disparities,[6] with male athletes receiving more attention and resources in research funding and media coverage.[12] Though this has changed moderately in the past three decades, the consequences of this bias are significant, as it limits our understanding of how concussions may manifest differently in female athletes and obscures potentially vital insights into sex-specific risk factors, symptoms, and recovery patterns[13]. Despite evidence that female athletes experience unique physiological and psychosocial responses to concussions, medical research has only recently started to recognize this differentiation.[1,3-5] A recent summary of published research results identified that female athletes need to be adequately represented in sports-related concussions literature. [1,6,7] In 2021, a systematic review[1] of 161 distinct studies with human participants was conducted to quantify female athlete participation in research. The literature revealed that most studies of sports-related concussions relied on samples that were 80.1% male. Furthermore, 40.4% of these studies did not include female participants. While male and female athletes share some similarities, they display various biomechanical, physiological, and neuroanatomical differences that can adversely affect concussion care and recovery.[8] So, while concussion research has seen marked progress, despite these strides, much is still to be learned about the effects of concussions on female athletes.

In what follows, we present a modest, non-exhaustive, scoping review having two particular goals: (i) to examine current and previous literature on neuropathological findings and gaps in female-concussed populations, (ii) to highlight the potential utility of neuroinformatics approaches to effectively characterize, measure, and model the various complexities of sports-related concussion, emphasizing the disparities present in current concussion research as they pertain the female athletes. Furthermore, this appraisal accentuates the need for further advances in neuroinformatics (e.g., multimodal neuroimaging experimental design, data science approaches, etc.) applied concerning female athlete populations, their interpretability, and capability for mitigating sources of assessment and treatment bias. Improved clinical decision making will require both that data collection focuses on female athletes and for analytical techniques to be interpretable in order to link research findings to underlying mechanisms.

*Increasing Recognition of Female Sports Concussions*

To address this disparity, national governing bodies such as the National Collegiate Athletic Association (NCAA) and the Department of Defense (DoD)[14] have collaborated with research experts and universities to conduct research that informs injury protocols and has begun to bridge the gap between males and females through the CARE Consortium.[14] One of the main goals of the CARE Consortium was to identify sex-specific factors that aid researchers and healthcare professionals in developing injury prevention and management strategies that are more effective with the hope of developing future strategies that can be tailored to meet each athlete's unique needs.[14] Beyond the NCAA and DoD, other public sector organizations are ardent about promoting equality in concussion research. For example, PINK Concussions,[15] is a non-profit group dedicated to educating females about brain injuries like sports-related concussions. PINK has been an advocate in creating positive change and innovation by promoting gender-specific, evidence-based approaches for identifying, managing, and supporting women with brain injuries. Additionally, the Concussion Legacy Foundation[16] (CLF), another non-profit organization focusing on promoting female presentation within concussion research, disseminates current research on brain injuries at local, state, national, and international events, as well as through all forms of media.

*Hormonal Implications in Concussion Recovery*

There has been a long-standing debate in the field of concussion injury and management regarding the potential biological differences between genders that result in more severe post-injury outcomes.[87-94] However, it has been postulated that sex hormones may be a possible explanation. [96] A few recent reports[95,96]found that women who sustained a mild traumatic brain injury (MTBI) during the luteal phase of their menstrual cycle had a significantly lower quality of life and indicators of health 1 month after discharge than those females who were in the follicular phase of the cycle or were taking hormone contraceptives.[96] Other researchers have suggested that estrogen and progesterone-mediated neuroprotection is thought to be related to their effects on hormone receptors, direct antioxidant effects, effects on astrocytes and microglia, modulation of the inflammatory response to injury, and effects on mediating glutamate excitotoxicity, among others. [94,97-99] However, the true nature of the relationship between increase in estrogen and recovery length post-concussion is poorly understood.

*Chronic Traumatic Encephalopathy Female-Related Research*

Diagnosing Chronic Traumatic Encephalopathy (CTE) can be diagnosed within brain tissue post-mortem,[66,67] as this diagnosis requires neuropathologic evidence of perivascular hyperphosphorylated tau (p-tau) aggregates in neurons, with or without astrocytes, typically at the depths of the sulci in the cerebral cortex. [68-70] Neuroinformatics may offer a valuable toolkit for shedding light on the phenomenon of CTE in athletes, thought to result from exposed repetitive head impacts (RHI).[71-73] It is important to note CTE is a general term for the complex neurodegenerative condition associated with RHI,[67] including multiple concussions. [8, 74] RHI also does not directly always result in a diagnosable concussion, as RHI results come from impact-induced head accelerations that produce tensile and shear strains within the brain's tissues.[67] Moreover, repetitive sub-concussive head impact exposure has been correlated with reducing concussion tolerance.[65] Therefore, a sports-related concussion is defined as a singular injury, whereas RHI is exposure to brain shearing over some time.[65] Neuroinformatics allows for integrating diverse data sources, such as brain imaging, genetic information, and longitudinal clinical data, providing a holistic view of the factors contributing to CTE development. By applying advanced computational methods and ML, researchers would be better able to uncover subtle gender-specific patterns, temporal relationships, and hormonal-specific differences that may be essential in elucidating the pathophysiology of CTE.

Over 97% of CTE cases published have been reported in individuals with known exposure to repetitive head impacts (RHI), with the youngest being a 17-year-old soccer player.[119] There has been some discussion surrounding the potential causal relationship between sports-related RHI and CTE. Nevertheless, a new study makes a strong case for any alternative hypotheses that could account for the association between sports-related RHI and CTE. [75 ,76,118] An international brain autopsy study of women who had experienced intimate partner violence further reveals substantial damage in the brain, but no evidence of chronic traumatic encephalopathy (CTE), the neurodegenerative disease recognized among contact sports athletes who sustain repeated head trauma.[118] These findings become more prevalent when a 17-year-old football athlete exhibited symptoms of the disease after dying at a relatively young age.[75,77,119] However, the development of CTE in professional female athletes and the effects RHI has on the plasticity of the female brain needs further investigation.[75,78,119]

Despite confirmed diagnoses coming post-mortem, certain diagnostic features are present in over 70% of confirmed CTE cases, which can be classified into three categories: cognitive, behavioral, and mood.[71,79] One-third of the cases showed symptoms at retirement from the sport, while half showed symptoms within four years of stopping play. [80] Thus, a clinical criteria was created for the syndrome associated with CTE, known as traumatic encephalopathy syndrome (TES). [81,82] TES can be diagnosed based on criteria established by the National Institute of Neurological Disorders and Stroke (NINDS). [81] The main symptoms of TES include difficulties with thinking and memory and regulating emotions and behavior. People with TES may act impulsively or explosively and have trouble controlling their emotions.[81] The supportive features of CTE include delayed onset of core clinical features, parkinsonism, other motor signs (including amyotrophic lateral sclerosis), depression, anxiety, apathy, and paranoia. [75] These symptoms have been documented to be present years after RHI exposure has ceased.[75,79] The diagnosis of TES does not necessarily confirm the presence of CTE neuropathological changes, such as p-tau accumulation.[82] Rather, TES is a diagnosis of a clinical syndrome that is linked with a history of repetitive brain trauma,[82] clinically evaluated based on provisional diagnostic classifications of 'probable CTE,' 'possible CTE,' and 'unlikely CTE.'[82-85] As research on the clinical presentation of CTE is still in its early stages, establishing significant diagnostic criteria for "probable CTE" based purely on clinical features and course, like the criteria used for probable Alzheimer's Disease dementia in the National Institute on Aging-Alzheimer's Association (NIA-AA) AD diagnostic criteria, is not currently feasible.[86]

In leveraging the severity parameter within clinical presentations, controlling for gender could provide valuable insights to the provider and the patient. However, current studies into the pathology of CTE and TES clinical presentation include inadequate female participation or lack of female participation. For instance, one recent study included only one female out of six participants[82] whereas several others primarily studied TES by examining male amateur and professional boxers[83-85], including no female participants at all. A lack of female representation in sophisticated critical sports medicine studies could lead to misdiagnosis of TES in females, highlighting the significance of adequate female representation in research.

*More Data, More Questions, Greater Need for Data Science?*

While neuroimaging techniques constitute useful tools to evaluate *in vivo* brain structural (MRI and DTI), functional (fMRI, resting state-fMRI) and molecular (PET, SPECT) changes in TBI versus otherwise healthy brain, other data-intensive methods of monitoring brain function, like electroencephalography (EEG) and magnetoencephalography (MEG) are also being applied.[23,25,119] Whole genome-analytics are also being explored to identify gene expression patterns which might predict the differential rates of concussion recovery between males and females.[86,120] Coupled with electronic health records (EHR) research[121] and U.S.-based national resources like the Federal Interagency Traumatic Brain Injury Resource (FITBIR), [122] there is an increasing need for advanced analytics in order to expedite research and clinical treatment options. Emerging technologies such as ML and AI, in particular, have the potential to revolutionize the understanding of complex brain-health outcomes and AI-based diagnosis tools hold tremendous promise in dissecting the intricate relationship between post-concussion recovery in female athletes, their return to play, as well as their brain health in later life. In particular, implementing advanced ML and AI techniques may enable the evaluation of the comprehensive and integrated perspective of the lucid and consistent picture of the neuroplastic changes associated with hormonal fluctuations during the menstrual cycle. ML may be able to help identify objective biomarkers for concussion as well as useful in predicting concussion recovery. In leveraging these modeling approaches, it may be possible to gain a more nuanced understanding of the complex interplay between hormonal changes and the brain's ability to adapt, thereby paving the way for more personalized and effective treatments.[101]

Rapid increases in clinical and laboratory data collection have demanded a particular emphasis being placed on machine learning (ML) models and artificial intelligence (AI) for improving how neuroimaging data might be modeled and TBI cases classified.[25-27] Like many instances of mild traumatic brain injuries (mTBI), concussions do not show visible or specific signs of injury, such as bleeding, and no structural abnormalities are detected in brain imaging with either gender. [17,18,19,20, 22-24] ML can harness the ability to distinguish these small microstructural differences between two subgroups, as well as relate brain images to clinical or behavioral observations. Moreover, ML-supervised learning, algorithms based on labeled datasets to predict outcomes accurately, can help better understand the recovery process and identify differences in recovery patterns between male and female patients.[28,29] Meanwhile, ML unsupervised learning, which uses machine learning algorithms to analyze and cluster unlabeled data sets, can reveal hidden structures in sets of images or uncover sub-populations in large groups of patients.[30] By analyzing the microstructural and neurophysiological properties of the brain following a concussion, more personalized and effective treatment interventions could be possible, ultimately improving the quality of care provided to patients.[8,30-32]

In the last decade, using ML techniques is highly beneficial in differentiating gender-specific patterns in neuroimaging data.[102-104] In 2023, researchers from NYU Grossman School of Medicine showed for the first time that ML methods were capable of accurately distinguishing between the brains of male athletes who played contact sports like football versus non-contact sports like track and field.[103] Churchill *et al.* also employed ML to detect concussion recovery differences in male and female athletes. Their findings reported no difference in acute symptoms or recovery time, however, all neuroimaging measures showed significant sex differences during recovery. [104] Similarly, advanced deep learning models were 90% successful at determining whether fMRI scans of brain activity came from a female or a male brain. [102] These results highlight the utility of ML and AI methods to craft differential neurological biomarkers specific to gender and even sport that do not appear on conventional MRI scans.

Data science-based techniques such as neural networks (NNs) and statistical learning also have major advantages: first, detecting subtle changes, such as sex-related neuropathological differences, may require a particular combination of sensitive neuropsychological and neurological assessments; thus, employing more advanced statistical models is necessary.[32-35] Second is the ability to handle data with small female sample sizes, and female retention defects that are seen in current concussion research. Small sample sizes are common in sports-concussion research and a smaller portion are female. Therefore, ML can help to obtain sufficient results; high-quality within a small female sample size can prove better than a large sample of lower-quality data in the case of statistical ML.[36] Third, creating brain imaging pipelines [37] for analyzing brain imaging data in females, which funnels into a repository, can be a helpful source for investigators to collaborate.[34] For example, Leech and colleagues created a well-established algorithm for brain imaging studies in syndromes such as autism[47.] The algorithm estimates the performance of other nearby pipelines and datasets from similar subgroups to identify gender-specific clusters.[37] Moreover, data science approaches have sophisticated methods of handling missing data through imputation and *post-hoc* analyses to reduce false positive results in female athletes.[38,39]

ML has also been utilized to pinpoint biomarkers in other health-related outcomes to construct a predictive model that identified specific patterns within the data. Dabek and Caban[48] utilized a ML to develop a model that accurately predicted the likelihood of military service members developing posttraumatic stress disorder after a concussion, which was then validated. Notably, deviations in connectivity structures from a ML model predicted brain tumor recurrence up to two months in advance.[44,54] Similarly, Kampaki[48] conducted a preliminary classification study, leveraging ML tools to estimate injury recovery time for professional football (soccer) players. Kim[49] also successfully employed ML algorithms with high accuracy to predict patient outcomes with persistent post-concussive symptoms successfully. Falcone et al[50] also employed ML techniques based on vowel sounds extracted from speech recordings and successfully detected concussion incidents with high prediction accuracy.

With the use of ML, it is possible to extract meaningful features from images, which can then be used to generate clinically significant biomarkers specific to gender.[51] By leveraging advanced technologies to extract and analyze data from medical images, researchers and healthcare professionals can gain new insights into the underlying mechanisms of various diseases and develop more effective treatment strategies. [52,53] Neuroinformatics within the female brain represents a promising avenue for advancing the field of medicine and improving female patient outcomes.[31] As such, ML is a powerful tool in detecting and quantifying even the most subtle differences in brain activity between two subpopulations.[52-54]

Deep learning is a more powerful tool for analyzing brain images in ML, which can extract non-linear network structure, realize approximation of complex functions, characterize distributed representation of input data, and demonstrate the powerful ability to learn the essential features of datasets based on a small size of samples.[55] Differences in) with ML techniques, researchers have revealed subtle changes related to normal brain development.[56,57]

*A Role for Neuroinformatics?*

In many ways, the field of Neuroinformatics is <u>where</u> brain science meets data science – that is, the intersection of brain-related datatypes, measurements, models, and meta-data suitable for characterization by modern computational data science methods. One area where neuroinformatics has been particularly impactful is in the area of human neuroimaging studies identifying individual biomarkers and combinations of biomarkers, which aim to improve the accuracy of diagnosis of a sports-related concussion or even predict future neurodegeneration associated with repetitive head impacts. It has facilitated cross-modal integration has unveiled the relationships between brain structure and function, shedding light on the neural underpinnings of cognition, behavior, and neurological disorders. Furthermore, a focus on neuroinformatics has played a pivotal role in establishing large-scale, openly accessible databases like the Human Connectome Project (HCP), the Alzheimer's Disease Neuroimaging Initiative (ADNI), and others. These resources have promoted collaboration and knowledge sharing within the scientific community and accelerated the development of advanced analysis methods, such as network neuroscience, which provides insights into the brain's intricate connectivity patterns.[25, 45,46] Neuroinformatics has been a foremost catalyst for groundbreaking discoveries in human neuroimaging, leading to a deeper understanding of the brain's structure, function, and connectivity and offering new avenues for improving human brain health and addressing neurological challenges.

Neuroinformatics approaches have previously shown promise as valuable analytical tools in addressing the multifaceted challenges that arise from post-concussion neuroimaging research.[55,57-65] Nevertheless, existing studies are yet to provide consistent results on exploring the difference of brain structure between men and women.[57] In a recent study that included female participation, a deep convolutional neural network (CNN) used the convolution kernels to extract the features of the image by using the designed 3D PCNN algorithm. In line with previous work, this study confirmed that gender-related differences exist in the whole-brain fractional anisotropy (FA) images and in each specific brain region. [57-58] Due to limited research available, more extensive, and diverse datasets are needed to explore these neurological gender differences at a greater length. Continued research, collaboration between data science, clinical neurology, and sports medicine experts, and rigorous testing are essential to ensure that data science-based analytics meet gender-equitable standards and have a promising and reliable future in advancing our understanding and management of sports-related concussions.

One specific challenge concerns how computational methods like ML, deep learning, NNs, etc. are <u>explainable</u>. Model explainability involves describing to humans how and why a ML model's made a decision. [34] It should not be unreasonable that a human would like to comprehend an algorithm and its output, and by analyzing the decisions and results of ML/AI models, gain and understanding about the reasoning behind the system's decision.[40] This is especially important for 'black box' models, which learn directly from data without human guidance.[34,40]  ML models have a slight advantage of having sufficient explainability[40] compared to more deep-learning models.[38] But as such models become still more complicated, nuanced, and sophisticated, will this always remain true?

Practitioners of neuroinformatics, having armed themselves over decades with advanced analytic tools specific to a range of brain data types, are essential for making ML/AI methods applied to neuroscience data explainable, particularly needed in contexts such as sports-related concussions.  Explainable ML/AI methods aim to make the decision-making process of machine learning models transparent and interpretable to end-users, including researchers, clinicians, and athletes. Neuroinformatics provides a foundation for incorporating domain-specific knowledge into AI models, allowing for the generation of explanations that are not only accurate but also comprehensible to stakeholders with varying levels of expertise. This transparency fosters trust in AI-based diagnostic and prognostic tools for assessing sports-related concussions, facilitating their adoption in clinical practice and athletic settings.

What is more, neuroinformatics promotes rigorous validation and reproducibility of ML/AI algorithms through standardized evaluation protocols and benchmark datasets. By establishing common benchmarks for concussion diagnosis and outcome prediction, researchers can systematically compare different AI approaches and ensure their generalizability across diverse populations and experimental conditions. This systematic validation enhances the reliability and robustness of explainable ML/AI methods, reinforcing their utility in real-world applications.

Furthermore, within the small studies found to have been conducted on concussed female athletes that utilized ML in neuroimaging, these studies exhibited encouraging results in precisely and objectively evaluating gender-specific neuropathological characteristics.[41-42] However, further research is needed to investigate the neural pathological mechanisms involved in female concussions and comprehend the significance of sex differences following a concussion.[29] Although some promising insights have been discovered, more research using neuroimaging and ML is required to investigate sex differences after sports-related concussions fully.[43,44] Furthermore, advancements in imaging techniques may lead to better comprehension of brain injuries in females in the acute and chronic phases.

Achieving transparency in these models is an imperative for several reasons. First, it enables researchers and healthcare professionals to comprehend and interpret the intricate relationships and patterns of head injury exclusive to one's biological gender.[42] This understanding is essential for making informed decisions regarding concussion diagnosis, treatment, and prevention strategies that may vary between male and female athletes due to sex-specific differences. Moreover, explainable models enhance the credibility and trustworthiness of findings, making it easier to communicate results to stakeholders, athletes, and their families, ultimately contributing to more effective, tailored, and equitable care for athletes of all genders.

Through integrating diverse datasets, advanced analytical techniques, and ML/AI models, neuroinformatics allows for identifying nuanced sex-specific patterns and distinctions in concussion-related data. It enables researchers to delve into the underlying neural mechanisms, differential symptom presentations, and recovery pathways, which a historical male-centric focus in traditional research may overshadow. Elucidating the underlying mechanisms behind concussions require moving beyond traditional statistical methods not only because of the subtlety of differences, but also because of the limitations inherent in retrospective sport-related concussion populations, where injuries, mechanisms, time to data collection, and physiology of participants can be highly heterogenous and not always available. Moreover, neuroinformatics facilitates the creation of more balanced, inclusive datasets, ensuring equitable representation of brain injuries in male as well as in female athletes. On the other end of the analysis trajectory, neuroinformatics encompasses advanced statistical methodologies which incorporate variability along multiple dimensions, enabling the inclusion of a diverse array of sports, for example, where concussions may result from different impact types (i.e. from another player, playing surface, rotational/torque/shearing, or impacts from athletic equipment). This is an especially important consideration for research in female athletes since a great deal of concussion work has been performed in American football, a sport where concussive forces can be very different than those experienced in sports popular with female athletes. As a result, the approach typically embraced in neuroinformatics has put a premium on the accuracy and interpretability of findings. It has long sought to ensure that the insights gained from analysis of brain data types are universally applicable, make sense clinically, and are sound with respect to foundational neuroanatomy. Adopting such a mindset for dealing with data that affects male subjects differently from female subjects, will ultimately contribute to more equitable and effective concussion management strategy specifically tailored to each athlete, inclusive of the athlete's sex or gender.[24]

In summary, neuroinformatics provides a solid foundation for enhancing the explainability of ML/AI methods applied to neuroscience data, particularly in the context of sports-related concussions, and those factors which contribute to brain health in female athletes. By facilitating data integration, feature extraction, knowledge representation, model transparency, and validation, neuroinformaticists, with their vast knowledge and experience, are just the community who can help bring clarity, context, and explainability to the modeling needed for vexing problems like differentiating male and female sports-related head injuries, mapping their trajectories, and unambiguously defining what recovery looks like.

*Discussion*

As contact sports have traditionally been male-dominated, they have been recognized as being more prone to head injuries. However, research suggests that females may be more vulnerable to sports-related concussions,[3,7,106,107] even when participation rates are considered. Although the number of reported cases of CTE in females is currently limited, it is vital to consider the potential risks associated with sports participation for both genders, particularly considering the growing body of evidence on the subject.[75,77] CTE is a distinctive neurodegenerative disorder which is said to be triggered by repeated brain injury, frequently sustained in contact sports.[115-116] The condition manifests as a gradual decline in cognitive function and can be accompanied by early indications such as memory loss, cognitive impairment, depression, and impulse control issues. In severe instances, the disease's gradual progression can result in dementia-like symptoms[100-102] and, in extreme instances, self-harm.[75,77]

Due to sports-related concussions being a complex and multifaceted condition, often accompanied by physiological, sociocognitive, and psychological symptoms,[108,109] it is important to note that clinical recovery, as measured by symptom resolution, does not necessarily equate to complete neurobiological recovery.[1] As mentioned, one of the key benefits of neuroinformatics-based approaches, such as ML, is the ability to identify non-linear relationships and high-order interactions between multiple variables, a result traditional statistics cannot produce.[110] Considering a major limitation of identifying correlations between neurological disorders and concussions can be due to the large variance in the time elapsed since the last concussion and time of imaging, these models could obtain the ability to learn and interpret results despite the significant time lag.[28,50] With its advanced capabilities, ML has the potential to be the ideal solution for effectively addressing the intricacies of the neuropathological process of concussions in females.[27,111,112] Thus, employing a multifactorial ML approach that considers the temporal patterns of various fluctuating biomarkers and alterations in cognitive and other functional metrics may be a superior method for identifying persistent concussion-related impairments or detecting the early stages of a chronic condition stemming from repeated head impact occurrences. [113,114]

Current research in ML and sports-related concussion has primarily focused on male, adolescent, or adult athletes from team sports. [1,12] While some studies have included mixed-gender samples, most participants reported being male. Thus, the lack of research on female athletes in ML for sports-related concussion management highlights the gender imbalance in sports and exercise science research.[117] Although women are included in sports science research and concussion injury research, the number of female participants is significantly lower than that of male participants.[1,12,117] This disparity seems more pronounced in ML studies for concussion management than in the broader field of sports and exercise. Conducting additional studies utilizing diverse athlete data for analysis through ML algorithms may help bridge this critical knowledge gap and improve the prevention and treatment of sports-concussion, ultimately benefiting athlete well-being.

The examination of neuroimaging for concussed female athletes has revealed a significant lack of research regarding neuropathology and ML, resulting in a critical gap in statistical information and pathology.[118] This gap in research is particularly concerning given the increasing frequency of concussions in female athletes and the potential long-term health consequences that may arise from inadequate diagnosis and treatment. Addressing this issue requires immediate attention and resources to advance the field of neuroimaging and improve our understanding of the pathology of concussions in female athletes. The utility of neuroinformatics approaches here is important since risk of CTE in sports appears to be closely linked to an athlete's career duration and the frequency of brain injuries sustained.[105] To achieve this, it is essential to prioritize research utilizing neuroinformatics approaches to develop a comprehensive framework for research, defining what sports-concussion 'looks like in male as well as in female athletes, aimed at properly diagnosing and treating concussions in females.

*Conclusion*

Sports-related concussions constitute a multifaceted condition, characterized by a diverse array of physiological, cognitive, and psychological symptoms. Importantly, it's essential to recognize that the mere resolution of clinical symptoms does not necessarily imply a complete neurobiological recovery. In this complex landscape, the power of ML emerges as a beacon of hope, primarily due to its unique ability to unearth non-linear relationships and high-order interactions among various variables, a feat that traditional statistical methods cannot achieve. This capability opens the door to a deeper understanding of the intricate nuances within sports-related concussion, potentially revolutionizing the way we approach its management, diagnosis, and treatment.

However, a critical issue looms large in the domain of sports-related concussion research: a glaring gender imbalance. This gender disparity raises an urgent call to action to rectify this gap in sports-concussion research and clinical practice. To do so, the field of concussion research must embark on comprehensive research initiatives, rooted in foundational, evidence-based science, which encompass a diverse range of athlete data and employ neuroinformatics techniques to fully appreciate the intricacies of concussion and recovery in both male and female athletes. In essence, neuroinformatics is a critical tool in advancing our understanding of clinical syndromes stemming from sports-related concussions and can ultimately aid in the development of more effective prevention and intervention strategies.

This *call to arms* is not just a matter of academic concern; it holds profound implications for all athletes' well-being and long-term health outcomes. Concussion management, diagnosis, and treatment should be equitable and effective for all athletes, regardless of their sex or gender. Hence, a collective and concerted effort is imperative to harness the potential of neuroinformatics and advanced research methodologies to chart a path toward comprehensive, yet interpretable, sex- and gender-inclusive sports concussion studies.

**References**

1. D'Lauro, C., Jones, E. R., Swope, L. M., Anderson, M. N., Broglio, S., & Schmidt, J. D. (2022). Under-representation of female athletes in research informing influential concussion consensus and position statements: an evidence review and synthesis. British journal of sports medicine, bjsports-2021-105045. Advance online publication. https://doi.org/10.1136/bjsports-2021-105045

2. NIH is required by law (Public Health Service Act Section 492B, 42 U.S.C. Section 289a-2, added by Section 101 of Public Law 103-43 on June 10, 1993) to ensure that women and minorities are included in all clinical research as appropriate to the scientific question under study. The law further requires that the design of a Phase III clinical trial must enable investigators to conduct a valid analysis of the differences in the effect of an intervention by sex and race/ethnicity if pertinent.

3. Covassin, T., & Elbin, R. J. (2011). The Female Athlete: The Role of Gender in the Assessment and Management of Sport-Related Concussion. Clinics in Sports Medicine, 30(1), 125–131. https://doi.org/10.1016/j.csm.2010.08.001

4. McAllister, T., & McCrea, M. (2017). Long-Term Cognitive and Neuropsychiatric Consequences of Repetitive Concussion and Head-Impact Exposure. Journal of Athletic Training, 52(3), 309–317. https://doi.org/10.4085/1062-6050-52.1.14

5. Stone, S., Lee, B., Garrison, J. C., Blueitt, D., & Creed, K. (2017). Sex Differences in Time to Return-to-Play Progression After Sport-Related Concussion. Sports health, 9(1), 41–44. https://doi.org/10.1177/1941738116672184

6. Courtenay W. H. (2000). Constructions of masculinity and their influence on men's well-being: a theory of gender and health. Social science & medicine (1982), 50(10), 1385–1401. https://doi.org/10.1016/s0277-9536(99)00390-1Nowatzki  N , Grant  KR. Sex is not enough: the need for gender-based analysis in health research. Health Care Women Int2011;32:263–77.doi:10.1080/07399332.2010.519838

7. Covassin, T., Swanik, C. B., & Sachs, M. L. (2003). Epidemiological considerations of concussions among intercollegiate athletes. Applied neuropsychology, 10(1), 12–22. https://doi.org/10.1207/S15324826AN1001_3

8. Chamard, E., Lassonde, M., Henry, L., Tremblay, J., Boulanger, Y., De Beaumont, L., & Théoret, H. (2013). Neurometabolic and microstructural alterations following a sports-related concussion in female athletes. Brain Injury, 27(9), 1038–1046.

9. McCrory, P., Meeuwisse, W., Dvořák, J., Aubry, M., Bailes, J., Broglio, S., Cantu, R. C., Cassidy, D., Echemendia, R. J., Castellani, R. J., Davis, G. A., Ellenbogen, R., Emery, C., Engebretsen, L., Feddermann-Demont, N., Giza, C. C., Guskiewicz, K. M., Herring, S., Iverson, G. L., Johnston, K. M., … Vos, P. E. (2017). Consensus statement on concussion in sport-the 5th international conference on concussion in sport held in Berlin, October 2016. British journal of sports medicine, 51(11), 838–847. https://doi.org/10.1136/bjsports-2017-097699

10. Asken, B. M., McCrea, M. A., Clugston, J. R., Snyder, A. R., Houck, Z. M., & Bauer, R. M. (2016). "Playing Through It": Delayed Reporting and Removal From Athletic Activity After Concussion Predicts Prolonged Recovery. Journal of athletic training, 51(4), 329–335. https://doi.org/10.4085/1062-6050-51.5.02

11. Koerte, I. K., Schultz, V., Sydnor, V. J., Howell, D. R., Guenette, J. P., Dennis, E., Kochsiek, J., Kaufmann, D., Sollmann, N., Mondello, S., Shenton, M. E., & Lin, A. P. (2020). Sex-Related Differences in the Effects of Sports-Related Concussion: A Review. Journal of neuroimaging : official journal of the American Society of Neuroimaging, 30(4), 387–409. https://doi.org/10.1111/jon.12726

12. Cowley, E. S., Olenick, A. A., McNulty, K. L., & Ross, E. Z. (2021). "Invisible Sportswomen": The Sex Data Gap in Sport and Exercise Science Research. Women in Sport and Physical Activity Journal, 29(2), 146–151. https://doi.org/10.1123/wspaj.2021-0028

13. de Borja, C., Chang, C. J., Watkins, R., & Senter, C. (2022). Optimizing Health and Athletic Performance for Women. Current reviews in musculoskeletal medicine, 15(1), 10–20. https://doi.org/10.1007/s12178-021-09735-2

14. Broglio, S. P., McCrea, M., McAllister, T., Harezlak, J., Katz, B., Hack, D., Hainline, B., & CARE Consortium Investigators (2017). A National Study on the Effects of Concussion in Collegiate Athletes and US Military Service Academy Members: The NCAA-DoD Concussion Assessment, Research and Education (CARE) Consortium Structure and Methods. Sports medicine (Auckland, N.Z.), 47(7), 1437–1451. https://doi.org/10.1007/s40279-017-0707-1

15. Snedaker, K. P. (2015). Mission Statement. Pink Concussions. https://www.pinkconcussions.com/pageus

16. Concussion Legacy Foundation. (2007). Mission & History. Mission & History | Concussion Legacy Foundation. https://concussionfoundation.org/about/mission-history

17. Raichle M. E. (1998). Behind the scenes of functional brain imaging: a historical and physiological perspective. Proceedings of the National Academy of Sciences of the United States of America, 95(3), 765–772. https://doi.org/10.1073/pnas.95.3.765

18. P. McCrory, W. Meeuwisse, K. Johnston, J. Dvorak, M. Aubry, M. Molloy, R.Cantu. (2008) Consensus statement on concussion in sport - the Third International Conference on Concussion in Sport held in Zurich, November Phys. Sportsmed., 37 (2009), pp. 141-159

19. Rajkomar, J. Dean, I. Kohane Machine learning in medicine. N Engl J Med, 380 (2019), pp. 1347-1358, [10.1056/NEJMra1814259](10.1056/NEJMra1814259)

20. Davenport T., Kalakota R. (2019). The potential for artificial intelligence in healthcare. Future Healthc J. 6, 94–98. 10.7861/futurehosp.6-2-94

21. Tierney, R. T., Sitler, M. R., Swanik, C. B., Swanik, K. A., Higgins, M., & Torg, J. (2005). Gender differences in head-neck segment dynamic stabilization during head acceleration. Medicine and science in sports and exercise, 37(2), 272–279. https://doi.org/10.1249/01.mss.0000152734.47516.aa

22. Blennow K., Hardy J., Zetterberg H. (2012) The neuropathology and neurobiology of traumatic brain injury. Neuron.;76(5):886-899. DOI: 10.1016/j.neuron.2012.11.021. Diagnostics (Basel). 2022;12(3):740. DOI: 10.3390/diagnostics12030740

23. Mavroudis, I., Kazis, D., Chowdhury, R., Petridis, F., Costa, V., Balmus, I. M., Ciobica, A., Luca, A. C., Radu, I., Dobrin, R. P., & Baloyannis, S. (2022). Post-Concussion Syndrome and Chronic Traumatic Encephalopathy: Narrative Review on the Neuropathology, Neuroimaging and Fluid Biomarkers. Diagnostics (Basel, Switzerland), 12(3), 740. [https://doi.org/10.3390/diagnostics12030740](https://doi.org/10.3390/diagnostics12030740)

24. International Concussion Society 5 years ago. (2019). The neuropathology of concussion. Retrieved from https://www.concussion.org/news/neuropathology-of-concussion/

25. Mateos-Pérez, J. M., Dadar, M., Lacalle-Aurioles, M., Iturria-Medina, Y., Zeighami, Y., & Evans, A. C. (2018). Structural neuroimaging as clinical predictor: A review of machine learning applications. NeuroImage. Clinical, 20, 506–522. https://doi.org/10.1016/j.nicl.2018.08.019

26. Bajwa, J., Munir, U., Nori, A., & Williams, B. (2021). Artificial intelligence in healthcare: transforming the practice of medicine. Future healthcare journal, 8(2), e188–e194. https://doi.org/10.7861/fhj.2021-0095

27. Tamez-Peña, J., Rosella, P., Totterman, S., Schreyer, E., Gonzalez, P., Venkataraman, A., & Meyers, S. P. (2022). Post-concussive mTBI in Student Athletes: MRI Features and Machine Learning. Frontiers in neurology, 12, 734329. https://doi.org/10.3389/fneur.2021.734329

28. Bergeron, M. F., Landset, S., Maugans, T. A., Williams, V. B., Collins, C. L., Wasserman, E. B., & Khoshgoftaar, T. M. (2019). Machine Learning in Modeling High School Sport Concussion Symptom Resolve. Medicine and science in sports and exercise, 51(7), 1362–1371. https://doi.org/10.1249/MSS.0000000000001903

29. D'Lauro, C., Johnson, B. R., McGinty, G., Allred, C. D., Campbell, D. E., & Jackson, J. C. (2018). Reconsidering Return-to-Play Times: A Broader Perspective on Concussion Recovery. Orthopaedic Journal of Sports Medicine, 6(3), 232596711876085. https://doi.org/10.1177/2325967118760854

30. Abraham, A., Pedregosa, F., Eickenberg, M., Gervais, P., Mueller, A., Kossaifi, J., Gramfort, A., Thirion, B., & Varoquaux, G. (2014). Machine learning for neuroimaging with scikit-learn. Frontiers in neuroinformatics, 8, 14. https://doi.org/10.3389/fninf.2014.00014

31. Singh, N. M., Harrod, J. B., Subramanian, S., Robinson, M., Chang, K., Cetin-Karayumak, S., Dalca, A. V., Eickhoff, S., Fox, M., Franke, L., Golland, P., Haehn, D., Iglesias, J. E., O'Donnell, L. J., Ou, Y., Rathi, Y., Siddiqi, S. H., Sun, H., Westover, M. B., Whitfield-Gabrieli, S., … Gollub, R. L. (2022). How Machine Learning is Powering Neuroimaging to Improve Brain Health. Neuroinformatics, 20(4), 943–964. https://doi.org/10.1007/s12021-022-09572-9

32. Valera, E. M., Joseph, A. C., Snedaker, K., Breiding, M. J., Robertson, C. L., Colantonio, A., Levin, H., Pugh, M. J., Yurgelun-Todd, D., Mannix, R., Bazarian, J. J., Turtzo, L. C., Turkstra, L. S., Begg, L., Cummings, D. M., & Bellgowan, P. S. F. (2021). Understanding Traumatic Brain Injury in Females: A State-of-the-Art Summary and Future Directions. The Journal of head trauma rehabilitation, 36(1), E1–E17. https://doi.org/10.1097/HTR.0000000000000652

33. Jiarui L., Tetsuo S., Yukio H. (2021) Introduce structural equation modelling to machine learning problems for building an explainable and persuasive model, SICE Journal of Control, Measurement, and System Integration, 14:2, 67-79,DOI: 10.1080/18824889.2021.1894040

34. Castillo, D., (2023). Explainability in machine learning. Seldon. https://www.seldon.io/explainability-in-machine-learning

35. Dabek, F., & Caban, J. J. (2015). Leveraging Big Data to Model the Likelihood of Developing Psychological Conditions After a Concussion. Procedia Computer Science, 53, 265–273. https://doi.org/10.1016/j.procs.2015.07.303

36. Faraway J., Augustin N., (2018) When small data beats big data. Stat Probab Lett; 136: 142–145.

37. Deweerdt, S. (2022). Machine learning streamlines neuroimaging data analysis. Spectrum. https://doi.org/10.53053/VYIR9955

38. Kang H. (2013). The prevention and handling of the missing data. Korean journal of anesthesiology, 64(5), 402–406. https://doi.org/10.4097/kjae.2013.64.5.402

39. Sterne, J. A., White, I. R., Carlin, J. B., Spratt, M., Royston, P., Kenward, M. G., Wood, A. M., & Carpenter, J. R. (2009). Multiple imputation for missing data in epidemiological and clinical research: potential and pitfalls. BMJ (Clinical research ed.), 338, b2393. https://doi.org/10.1136/bmj.b2393

40. Samek W., Wiegand T., Müller K-R. (2017). Explainable artificial intelligence: understanding, visualizing and interpreting deep learning models. arXiv, preprint arXiv:1708.08296.

41. Fleck, D. E., Ernest, N., Asch, R., Adler, C. M., Cohen, K., Yuan, W., Kunkel, B., Krikorian, R., Wade, S. L., & Babcock, L. (2021). Predicting Post-Concussion Symptom Recovery in Adolescents Using a Novel Artificial Intelligence. Journal of neurotrauma, 38(7), 830–836. https://doi.org/10.1089/neu.2020.7018

42. Bergeron, M. F., Landset, S., Maugans, T. A., Williams, V. B., Collins, C. L., Wasserman, E. B., & Khoshgoftaar, T. M. (2019). Machine Learning in Modeling High School Sport Concussion Symptom

Resolve. Medicine and science in sports and exercise, 51(7), 1362–1371. https://doi.org/10.1249/MSS.0000000000001903

43. Singh, N. M., Harrod, J. B., Subramanian, S., Robinson, M., Chang, K., Cetin-Karayumak, S., Dalca, A. V., Eickhoff, S., Fox, M., Franke, L., Golland, P., Haehn, D., Iglesias, J. E., O'Donnell, L. J., Ou, Y., Rathi, Y., Siddiqi, S. H., Sun, H., Westover, M. B., Whitfield-Gabrieli, S., … Gollub, R. L. (2022). How Machine Learning is Powering Neuroimaging to Improve Brain Health. Neuroinformatics, 20(4), 943–964. https://doi.org/10.1007/s12021-022-09572-9

44. Nenning, K. H., & Langs, G. (2022). Machine learning in neuroimaging: from research to clinical practice. Maschinelles Lernen in der Neurobildgebung: von der Forschung in die klinische Praxis. Radiologie (Heidelberg, Germany), 62(Suppl 1), 1–10. https://doi.org/10.1007/s00117-022-01051-1

45. Avberšek, L. K., & Repovš, G. (2022). Deep learning in neuroimaging data analysis: Applications, challenges, and solutions. Frontiers in neuroimaging, 1, 981642. https://doi.org/10.3389/fnimg.2022.981642

46. Vieira S., Pinaya W. H., Mechelli A. (2017). Using deep learning to investigate the neuroimaging correlates of psychiatric and neurological disorders: methods and applications. Neurosci. Biobehav. Rev. 74, 58–75. 10.1016/j.neubiorev.2017.01.002

47. M. Falcone, N. Yadav, C. Poellabauer and P. Flynn, Using isolated vowel sounds for classification of Mild Traumatic Brain Injury, 2013 IEEE International Conference on Acoustics, Speech and Signal Processing, Vancouver, BC, Canada, (2013), pp. 7577-7581, doi: 10.1109/ICASSP.2013.6639136.

48. Kampakis S. Predictive modelling of football injuries. 2016; Available from: http://arxiv.org/abs/1609.07480.

49. Kim, M. (2021). Predicting Post-Concussion Syndrome Outcomes with Machine Learning. https://doi.org/10.48550/ARXIV.2108.02570

50. Rosenblatt, C. K., Harriss, A., Babul, A. N., & Rosenblatt, S. A. (2021). Machine Learning for Subtyping Concussion Using a Clustering Approach. Frontiers in human neuroscience, 15, 716643. https://doi.org/10.3389/fnhum.2021.716643

51. Furtner, J., Berghoff, A. S., Albtoush, O. M., Woitek, R., Asenbaum, U., Prayer, D., Widhalm, G., Gatterbauer, B., Dieckmann, K., Birner, P., Aretin, B., Bartsch, R., Zielinski, C. C., Schöpf, V., & Preusser, M. (2017). Survival prediction using temporal muscle thickness measurements on cranial magnetic resonance images in patients with newly diagnosed brain metastases. European radiology, 27(8), 3167–3173. https://doi.org/10.1007/s00330-016-4707-6

52. Gray, K. R., Aljabar, P., Heckemann, R. A., Hammers, A., Rueckert, D., & Alzheimer's Disease Neuroimaging Initiative (2013). Random forest-based similarity measures for multi-modal classification of Alzheimer's disease. NeuroImage, 65, 167–175. https://doi.org/10.1016/j.neuroimage.2012.09.065

53. Fox, M. D., & Greicius, M. (2010). Clinical applications of resting state functional connectivity. Frontiers in systems neuroscience, 4, 19. https://doi.org/10.3389/fnsys.2010.00019

54. Nenning, K. H., Furtner, J., Kiesel, B., Schwartz, E., Roetzer, T., Fortelny, N., Bock, C., Grisold, A., Marko, M., Leutmezer, F., Liu, H., Golland, P., Stoecklein, S., Hainfellner, J. A., Kasprian, G., Prayer, D., Marosi, C., Widhalm, G., Woehrer, A., & Langs, G. (2020). Distributed changes of the functional connectome in patients with glioblastoma. Scientific reports, 10(1), 18312. https://doi.org/10.1038/s41598-020-74726-1

55. Tian R., Yang Y., Van Der Helm FCT Dewald JPA. (2018). A novel approach for modeling neural responses to joint perturbations using the NARMAX method and a hierarchical neural network. Front. Comput. Neurosci. 12:96. 10.3389/fncom.2018.00096

56. Lasi S., Szczepankiewicz F., Eriksson S., Nilsson M., Topgaard D. (2014). Microanisotropy imaging: quantification of microscopic diffusion anisotropy and orientational order parameter by diffusion MRI with magic-angle spinning of the q-vector. Front. Phys. 2:11 10.3389/fphy.2014.00011

57. Zeng N., Wang Z., Zhang H., Liu W., Alsaadi F. E. (2016). Deep belief networks for quantitative analysis of a gold immunochromatographic strip. Cogn. Comput. 8, 684–692. 10.1007/s12559-016-9404-x

58. Glickstein M., Doron K. (2008). Cerebellum: connections and functions. Cerebellum 7, 589–594. 10.1007/s12311-008-0074-4

59. Abe O., Aoki S., Hayashi N., Yamada H., Kunimatsu A., Mori H., et al.. (2002). Normal aging in the central nervous system: quantitative MR diffusion-tensor analysis. Neurobiol. Aging23, 433–441. 10.1016/S0197-4580(01)00318-9

60. Prendergast D. M., Ardekani B., Ikuta T., John M., Peters B., Derosse P., et al.. (2015). Age and sex effects on corpus callosum morphology across the lifespan. Human Brain Mapp. 36, 2691–2702. 10.1002/hbm.22800

61. Sullivan E. V., Rosenbloom M. J., Desmond J. E., Pfefferbaum A. (2001). Sex differences in corpus callosum size: relationship to age and intracranial size. Neurobiol. Aging 22, 603–611. 10.1016/S0197-4580(01)00232-9

62. Shirao N., Okamoto Y., Okada G., Ueda K., Yamawaki S. (2005). Gender differences in brain activity toward unpleasant linguistic stimuli concerning interpersonal relationships: an fMRI study. Eur. Arch. Psychiatr. Clin. 255, 327–333. 10.1007/s00406-005-0566-x

63. Lee J. Y., Kondziolka D. (2005). Thalamic deep brain stimulation for management of essential tremor. J. Neurosurgery 103, 400–403. 10.3171/jns.2005.103.3.0400  https://doi.org/10.1016/B978-0-12-823036-7.00023-2

64. Dvorak J, McCrory P, Kirkendall DT. Head injuries in the female football player: incidence, mechanisms, risk factors and management. Br J Sports Med. (2007) 41 (Suppl 1):i44–6. 10.1136/bjsm.2007.037960 [

65. Lakhan, S. E., & Kirchgessner, A. (2012). Chronic traumatic encephalopathy: the dangers of getting "dinged". SpringerPlus, 1, 2. https://doi.org/10.1186/2193-1801-1-2

66. Perrine K., Helcer J., Tsiouris A.J., Pisapia D.J., Stieg P. The Current Status of Research on Chronic Traumatic Encephalopathy. World Neurosurg. 2017;102:533–544. doi: 10.1016/j.wneu.2017.02.084

67. Van Horn, J. D., Bhattrai, A., & Irimia, A. (2017). Multimodal Imaging of Neurometabolic Pathology due to Traumatic Brain Injury. Trends in neurosciences, 40(1), 39–59. https://doi.org/10.1016/j.tins.2016.10.007

68. McKee AC, Cairns NJ, Dickson DW, et al.; TBI/CTE group . The first NINDS/NIBIB consensus meeting to define neuropathological criteria for the diagnosis of chronic traumatic encephalopathy. Acta Neuropathol. 2016;131(1):75-86. doi: 10.1007/s00401-015-1515-z

69. Bieniek KF, Cairns NJ, Crary JF, et al.; TBI/CTE Research Group . The Second NINDS/NIBIB Consensus Meeting to Define Neuropathological Criteria for the Diagnosis of Chronic Traumatic Encephalopathy. J Neuropathol Exp Neurol. 2021;80(3):210-219. doi: 10.1093/jnen/nlab001

70. Montenigro P.H., Corp D.T., Stein T.D., Cantu R.C., Stern R.A. Chronic traumatic encephalopathy: Historical origins and current perspective. Annu. Rev. Clin. Psychol. 2015;11:309–330. doi: 10.1146/annurev-clinpsy-032814-112814.

71. McKee A.C., Cantu R.C., Nowinski C.J., Hedley-Whyte E.T., Gavett B.E., Budson A.E., Santini V.E., Lee H.-S., Kubilus C.A., Stern R.A. Chronic traumatic encephalopathy in athletes: Progressive tauopathy after repetitive head injury. J. Neuropathol. Exp. Neurol. 2009;68:709–735. doi: 10.1097/NEN.0b013e3181a9d503.

72. Inserra C.J., DeVrieze B.W. Chronic Traumatic Encephalopathy. StatPearls Publishing LLC; Treasure Island, FL, USA: 2021.

73. Ling H., Neal J.W., Revesz T. Evolving concepts of chronic traumatic encephalopathy as a neuropathological entity. Neuropathol. Appl. Neurobiol. 2017;43:467–476. doi: 10.1111/nan.12425.

74. McKee, A. C., Mez, J., Abdolmohammadi, B., Butler, M., Huber, B. R., Uretsky, M., Babcock, K., Cherry, J. D., Alvarez, V. E., Martin, B., Tripodis, Y., Palmisano, J. N., Cormier, K. A., Kubilus, C. A., Nicks, R., Kirsch, D., Mahar, I., McHale, L., Nowinski, C., Cantu, R. C., … Alosco, M. L. (2023). Neuropathologic and Clinical Findings in Young Contact Sport Athletes Exposed to Repetitive Head Impacts. JAMA neurology, 80(10), 1037–1050. https://doi.org/10.1001/jamaneurol.2023.2907

75. Mez J, Daneshvar DH, Kiernan PT, et al.. Clinicopathological evaluation of chronic traumatic encephalopathy in players of American football. JAMA. 2017;318(4):360-370. doi: 10.1001/jama.2017.833

76. Suter, C. M., Affleck, A. J., Pearce, A. J., Junckerstorff, R., Lee, M., & Buckland, M. E. (2023). Chronic traumatic encephalopathy in a female ex-professional Australian rules footballer. Acta neuropathologica, 146(3), 547–549. https://doi.org/10.1007/s00401-023-02610-z

77. Bieniek, K. F., Blessing, M. M., Heckman, M. G., Diehl, N. N., Serie, A. M., Paolini, M. A., 2nd, Boeve, B. F., Savica, R., Reichard, R. R., & Dickson, D. W. (2020). Association between contact sports participation and chronic traumatic encephalopathy: a retrospective cohort study. Brain pathology (Zurich, Switzerland), 30(1), 63–74. https://doi.org/10.1111/bpa.12757

78. Jordan B.D. The clinical spectrum of sport-related traumatic brain injury. Nat. Rev. Neurol. 2013;9:222–230. doi: 10.1038/nrneurol.2013.33.

79. Chen L. What triggers tauopathy in chronic traumatic encephalopathy? Neural Regen. Res. 2018;13:985–986. doi: 10.4103/1673-5374.233439.

80. Katz DI, Bernick C, Dodick DW, et al.. National Institute of Neurological Disorders and Stroke consensus diagnostic criteria for traumatic encephalopathy syndrome. Neurology. 2021;96(18):848-863. doi:

81. Montenigro PH, Baugh CM, Daneshvar DH, et al.. Clinical subtypes of chronic traumatic encephalopathy: literature review and proposed research diagnostic criteria for traumatic encephalopathy syndrome. Alzheimers Res Ther. 2014;6(5):68. doi: 10.1186/s13195-014-0068-z

82. Jordan BD. The clinical spectrum of sport-related traumatic brain injury. Nat Rev Neurol. 2013;9:222–230. doi: 10.1038/nrneurol.2013.33.

83. Jordan BD. In: Medical Aspects of Boxing. 1. Jordan BD, editor. CRC Press Inc, Boca Raton; 1993. Chronic neurologic injuries in boxing; pp. 177–185.

84. Jordan BD. In: Neurobiology of Primary Dementia. Folstein MF, editor. American Psychiatric Press, Washington, DC; 1998. Dementia pugilistica; p. 191

85. Jack CR Jr, Albert MS, Knopman DS, McKhann GM, Sperling RA, Carrillo MC, Thies B, Phelps CH. Introduction to the recommendations from the National Institute on Aging-Alzheimer's Association workgroups on diagnostic guidelines for Alzheimer's disease. Alzheimers Dement. 2011;7:257–262. doi: 10.1016/j.jalz.2011.03.004

86. Covassin T, Moran R, Elbin RJ. Sex differences in reported concussion injury rates and time loss from participation: an update of the national collegiate athletic association injury surveillance program from 2004-2005 through 2008-2009. J Athl Train. (2016) 51:189–94. 10.4085/1062-6050-51.3.05

87. Covassin T, Schatz P, Swanik CB. Sex differences in neuropsychological function and post-concussion symptoms of concussed collegiate athletes. Neurosurgery. (2007) 61:345–50; discussion 350–1. 10.1227/01.NEU.0000279972.95060.CB

88. McDevitt, J., & Krynetskiy, E. (2017). Genetic findings in sport-related concussions: potential for individualized medicine?. *Concussion (London, England)*, *2*(1), CNC26. https://doi.org/10.2217/cnc-2016-0020

89. Kerr ZY, Roos KG, Djoko A, Dalton SL, Broglio SP, Marshall SW, et al.. Epidemiologic measures for quantifying the incidence of concussion in national collegiate athletic association sports. J Athl Train. (2017) 52:167–174. 10.4085/1062-6050-51.6.05

90. Mollayeva T, El-Khechen-Richandi G, Colantonio A. Sex & gender considerations in concussion research. Concussion.(2018) 3:CNC51. 10.2217/cnc-2017-0015

91. O'Connor KL, Baker MM, Dalton SL, Dompier TP, Broglio SP, Kerr ZY. Epidemiology of sport-related concussions in high school athletes: national athletic treatment, injury and outcomes network (NATION), 2011-2012 through 2013-2014. J Athl Train. (2017) 52:175–85. 10.4085/1062-6050-52.1.15

92. Gallagher V, Kramer N, Abbott K, Alexander J, Breiter H, Herrold A, et al.. The effects of sex differences and hormonal contraception on outcomes after collegiate sports-related concussion. J Neurotrauma. (2018) 35:1242–7. 10.1089/neu.2017.5453

93. Dubol, M., Epperson, C. N., Sacher, J., Pletzer, B., Derntl, B., Lanzenberger, R., Sundström-Poromaa, I., & Comasco, E. (2021). Neuroimaging the menstrual cycle: A multimodal systematic review. Frontiers in Neuroendocrinology, 60, 100878. https://doi.org/10.1016/j.yfrne.2020.100878

94. Wunderle K, Hoeger KM, Wasserman E, Bazarian JJ. Menstrual phase as predictor of outcome after mild traumatic brain injury in women. J Head Trauma Rehabil. (2014) 29:E1–8. 10.1097/HTR.0000000000000006

95. Brotfain E, Gruenbaum SE, Boyko M, Kutz R, Zlotnik A, Klein M. Neuroprotection by estrogen and progesterone in traumatic brain injury and spinal

96. Correia SC, Santos RX, Cardoso S, Carvalho C, Santos MS, Oliveira CR, et al.. Effects of estrogen in the brain: is it a neuroprotective agent in Alzheimer's disease? Curr Aging Sci. (2010) 3:113–26. 10.2174/1874609811003020113

97. Green PS, Simpkins JW. Neuroprotective effects of estrogens: potential mechanisms of action. Int J Dev Neurosci. (2000) 18:347–58. 10.1016/S0736-5748(00)00017-4

98. La Fountaine, M. F., Hill-Lombardi, V., Hohn, A. N., Leahy, C. L., & Testa, A. J. (2019). Preliminary Evidence for a Window of Increased Vulnerability to Sustain a Concussion in Females: A Brief Report. Frontiers in neurology, 10, 691. https://doi.org/10.3389/fneur.2019.00691

99. Gubbi, S., Hamet, P., Tremblay, J., Koch, C. A., & Hannah-Shmouni, F. (2019). Artificial Intelligence and Machine Learning in Endocrinology and Metabolism: The Dawn of a New Era. Frontiers in Endocrinology, 10, 185. https://doi.org/10.3389/fendo.2019.00185

100. Fu, C.H.Y., Erus, G., Fan, Y. et al. AI-based dimensional neuroimaging system for characterizing heterogeneity in brain structure and function in major depressive disorder: COORDINATE-MDD consortium design and rationale. BMC Psychiatry 23, 59 (2023). https://doi.org/10.1186/s12888-022-04509-7

101. Ryali, S.*, Zhang, Y,*, de los Angeles, C., Supekar, K. & Menon, V. (In Press). Deep learning models reveal replicable, generalizable, and behaviorally relevant sex differences in human functional brain organization. *Proceedings of the National Academy of Sciences USA*.

102. Chen, J., Chung, S., Li, T., Fieremans, E., Novikov, D. S., Wang, Y., & Lui, Y. W. (2023). Identifying relevant diffusion MRI microstructure biomarkers relating to exposure to repeated head impacts in contact sport athletes. The neuroradiology journal, 36(6), 693–701. https://doi.org/10.1177/19714009231177396

103. Churchill, N. W., Hutchison, M. G., Graham, S. J., & Schweizer, T. A. (2021). Sex differences in acute and long-term brain recovery after concussion. Human brain mapping, 42(18), 5814–5826. https://doi.org/10.1002/hbm.25591

104. Manley, G., Gardner, A. J., Schneider, K. J., Guskiewicz, K. M., Bailes, J., Cantu, R. C., Castellani, R. J., Turner, M., Jordan, B. D., Randolph, C., Dvořák, J., Hayden, K. A., Tator, C. H., McCrory, P., &

Iverson, G. L. (2017). A systematic review of potential long-term effects of sport-related concussion. British journal of sports medicine, 51(12), 969–977. https://doi.org/10.1136/bjsports-2017-097791

105.     Stone, S., Lee, B., Garrison, J. C., Blueitt, D., & Creed, K. (2017). Sex Differences in Time to Return-to-Play Progression After Sport-Related Concussion. Sports health, 9(1), 41–44. https://doi.org/10.1177/1941738116672184

106.     O'Connor, K. L., Baker, M. M., Dalton, S. L., Dompier, T. P., Broglio, S. P., & Kerr, Z. Y. (2017). Epidemiology of Sport-Related Concussions in High School Athletes: National Athletic Treatment, Injury and Outcomes Network (NATION), 2011–2012 Through 2013–2014. Journal of Athletic Training, 52(3), 175–185. https://doi.org/10.4085/1062-6050-52.1.15

107.     Covassin, T., & Elbin, R. J. (2010). The cognitive effects and decrements following concussion. Open access journal of sports medicine, 1, 55–61. https://doi.org/10.2147/oajsm.s6919

108.     Furtner, J., Berghoff, A. S., Albtoush, O. M., Woitek, R., Asenbaum, U., Prayer, D., Widhalm, G., Gatterbauer, B., Dieckmann, K., Birner, P., Aretin, B., Bartsch, R., Zielinski, C. C., Schöpf, V., & Preusser, M. (2017). Survival prediction using temporal muscle thickness measurements on cranial magnetic resonance images in patients with newly diagnosed brain metastases. European radiology, 27(8), 3167–3173. https://doi.org/10.1007/s00330-016-4707-6

109.     Mateos-Pérez, J. M., Dadar, M., Lacalle-Aurioles, M., Iturria-Medina, Y., Zeighami, Y., & Evans, A. C. (2018). Structural neuroimaging as clinical predictor: A review of machine learning applications. NeuroImage. Clinical, 20, 506–522. https://doi.org/10.1016/j.nicl.2018.08.019

110.     Bazarian, J. J., Elbin, R. J., Casa, D. J., Hotz, G. A., Neville, C., Lopez, R. M., Schnyer, D. M., Yeargin, S., & Covassin, T. (2021). Validation of a Machine Learning Brain Electrical Activity-Based Index to Aid in Diagnosing Concussion Among Athletes. JAMA network open, 4(2), e2037349. https://doi.org/10.1001/jamanetworkopen.2020.37349

111.     Jacob, D., Unnsteinsdóttir Kristensen, I. S., Aubonnet, R., Recenti, M., Donisi, L., Ricciardi, C., Svansson, H. Á. R., Agnarsdóttir, S., Colacino, A., Jónsdóttir, M. K., Kristjánsdóttir, H., Sigurjónsdóttir, H. Á., Cesarelli, M., Eggertsdóttir Claessen, L. Ó., Hassan, M., Petersen, H., & Gargiulo, P. (2022).

Towards defining biomarkers to evaluate concussions using virtual reality and a moving platform (BioVRSea). Scientific reports, 12(1), 8996. https://doi.org/10.1038/s41598-022-12822-0

112.    Fakhran, S., Yaeger, K., Collins, M., & Alhilali, L. (2014). Sex differences in white matter abnormalities after mild traumatic brain injury: localization and correlation with outcome. Radiology, 272(3), 815–823. https://doi.org/10.1148/radiol.14132512

113.    Smith D.H., Johnson V.E., Trojanowski J.Q., Stewart W. Chronic traumatic encephalopathy—Confusion and controversies. Nat. Rev. Neurol. 2019;15:179–183. doi: 10.1038/s41582-018-0114-8.

114.    Chen L. What triggers tauopathy in chronic traumatic encephalopathy? Neural Regen. Res. 2018;13:985–986. doi: 10.4103/1673-5374.233439.

115.    Malcolm D. (2023). Some problems of research exploring sex differences in sport-related concussions: a narrative review. Research in sports medicine (Print), 1–10. Advance online publication. https://doi.org/10.1080/15438627.2023.2271604

116.    Edelstein, R., & Van Horn, J. D. (2023). Modulating Factors Affecting Sports-Related Concussion Exposures: A Systematic Review and Analysis [Preprint]. Epidemiology. https://doi.org/10.1101/2023.03.08.23286974

117.    Dams-O'Connor, K., Seifert, A. C., Crary, J. F., Delman, B. N., Del Bigio, M. R., Kovacs, G. G., Lee, E. B., Nolan, A. L., Pruyser, A., Selmanovic, E., Stewart, W., Woodoff-Leith, E., & Folkerth, R. D. (2023). The neuropathology of intimate partner violence. Acta neuropathologica, 146(6), 803–815. https://doi.org/10.1007/s00401-023-02646-1

118.    McKee, A. C., Stein, T. D., Huber, B. R., Crary, J. F., Bieniek, K., Dickson, D., Alvarez, V. E., Cherry, J. D., Farrell, K., Butler, M., Uretsky, M., Abdolmohammadi, B., Alosco, M. L., Tripodis, Y., Mez, J., & Daneshvar, D. H. (2023). Chronic traumatic encephalopathy (CTE): criteria for neuropathological diagnosis and relationship to repetitive head impacts. Acta neuropathologica, 145(4), 371–394. https://doi.org/10.1007/s00401-023-02540-w

119.    Amyot, F., Arciniegas, D. B., Brazaitis, M. P., Curley, K. C., Diaz-Arrastia, R., Gandjbakhche, A., Herscovitch, P., Hinds, S. R., 2nd, Manley, G. T., Pacifico, A., Razumovsky, A., Riley, J., Salzer, W., Shih, R., Smirniotopoulos, J. G., & Stocker, D. (2015). A Review of the Effectiveness of Neuroimaging

Modalities for the Detection of Traumatic Brain Injury. *Journal of neurotrauma*, *32*(22), 1693–1721. https://doi.org/10.1089/neu.2013.3306

120.     Rauchman, S. H., Albert, J., Pinkhasov, A., & Reiss, A. B. (2022). Mild-to-Moderate Traumatic Brain Injury: A Review with Focus on the Visual System. *Neurology international*, *14*(2), 453–470. https://doi.org/10.3390/neurolint14020038

121.     Tian, T. Y., Zlateva, I., & Anderson, D. R. (2013). Using electronic health records data to identify patients with chronic pain in a primary care setting. *Journal of the American Medical Informatics Association : JAMIA*, *20*(e2), e275–e280. https://doi.org/10.1136/amiajnl-2013-001856

122.     U.S. Department of Health and Human Services. *Federal Interagency of Traumatic Brain Injury Research*. National Institutes of Health. https://fitbir.nih.gov/